%% file: output.tex
\documentclass{article}

\usepackage{subcaption} 

\usepackage{microtype}
\usepackage{graphicx}
\usepackage{booktabs} 

\usepackage{hyperref}

\usepackage[preprint]{icml2025}

\usepackage{amsmath}
\usepackage{amssymb}
\usepackage{mathtools}
\usepackage{amsthm}

\usepackage[capitalize,noabbrev]{cleveref}

\theoremstyle{plain}

\theoremstyle{definition}

\theoremstyle{remark}

\usepackage[textsize=tiny]{todonotes}

\usepackage{graphicx}
\usepackage{amsmath}
\usepackage{multirow}

\def\method{O1-Pruner}

\icmltitlerunning{O1-Pruner: Length-Harmonizing Fine-Tuning for O1-Like Reasoning Pruning}

\begin{document}

\twocolumn[
\icmltitle{O1-Pruner: Length-Harmonizing Fine-Tuning for O1-Like Reasoning Pruning}



\icmlsetsymbol{equal}{*}

\begin{icmlauthorlist}
\icmlauthor{Haotian Luo}{sysu}
\icmlauthor{Li Shen}{sysu} 
\icmlauthor{Haiying He}{cau}
\icmlauthor{Yibo Wang}{tsu}
\icmlauthor{Shiwei Liu}{oxf}\\
\icmlauthor{Wei Li}{didi}
\icmlauthor{Naiqiang Tan}{didi}
\icmlauthor{Xiaochun Cao}{sysu} 
\icmlauthor{Dacheng Tao}{ntu}

\end{icmlauthorlist}

\icmlaffiliation{cau}{China Agriculture University}
\icmlaffiliation{tsu}{Tsinghua University}
\icmlaffiliation{oxf}{University of Oxford}
\icmlaffiliation{didi}{Didichuxing Co. Ltd} 
\icmlaffiliation{sysu}{Shenzhen Campus of Sun Yat-sen University}
\icmlaffiliation{ntu}{Nanyang Technological University}


\icmlcorrespondingauthor{Li Shen}{mathshenli@gmail.com}

\icmlkeywords{Machine Learning, ICML}

\vskip 0.3in
]



\printAffiliationsAndNotice{}  

\input{text/abstract}
\input{text/introduction}
\input{text/related_work}
\input{text/rethinking}

\input{text/method}
\input{text/experiments}
\input{text/further_evaluation}


\section{Conclusion}
In this paper, we conducted simple experiments to validate the phenomenon of length disharmony in long-thought models during reasoning, which leads to redundant computational overhead in the inference phase. To address this issue, we formulated it as an optimization problem and proposed Length Harmonizing Fine-Tuning (\textbf{\method{}}) as a solution to optimize the model. Extensive experiments demonstrate that \textbf{\method{}} significantly reduces the length of the solutions generated by the model while achieves a modest improvement in accuracy, thereby substantially enabling more efficient reasoning. Additionally, we performed an in-depth analysis, including experiments on key hyperparameter and datasets of varying difficulty, to better understand the characteristics of \textbf{\method{}}.
\bibliography{deeplearningreference,reasoning}
\bibliographystyle{icml2025}




\end{document}

%% file: text/abstract.tex
\begin{abstract}
Recently, long-thought reasoning LLMs, such as OpenAI's O1, adopt extended reasoning processes similar to how humans ponder over complex problems. This reasoning paradigm significantly enhances the model's problem-solving abilities and achieves promising results. However, long-thought reasoning process leads to a substantial increase in inference time. A pressing challenge is reducing the inference overhead of long-thought LLMs while ensuring accuracy. 
In this paper, we identify that long-thought reasoning models struggle to effectively allocate token budgets based on problem difficulty and reasoning redundancies. To address this, we propose Length-Harmonizing Fine-Tuning (\method{}), aiming at minimizing reasoning overhead while maintaining accuracy. This effective fine-tuning method first estimates the LLM's baseline performance through pre-sampling and then uses RL-style fine-tuning to encourage the model to generate shorter reasoning processes under accuracy constraints. This allows the model to achieve efficient reasoning with lower redundancy while maintaining accuracy. Experiments on various mathematical reasoning benchmarks show that \method{} not only significantly reduces inference overhead but also achieves higher accuracy, providing a novel and promising solution to this challenge. Our code is coming soon at \href{https://github.com/StarDewXXX/O1-Pruner}{\textcolor{blue}{https://github.com/StarDewXXX/O1-Pruner}}
\end{abstract}


%% file: text/introduction.tex
\section{Introduction}
Reasoning represents a fundamental capability of large language models (LLMs), serving as a cornerstone in the advancement of artificial intelligence research \cite{huang-chang-2023-towards}. Recently OpenAI's O1\cite{o12024} have introduced long-thought reasoning models that mimic human-like problem-solving processes. In addition to O1, researchers have also developed models that inference with a similar long-thought reasoning pattern, such as Deepseek-R1 \cite{deepseek2024r1lite}, QwQ \cite{qwq-32b-preview} and Marco-o1\cite{zhao2024marcoo1openreasoningmodels}. These models leverage a long chain-of-thought framework, enabling them to tackle complex problems by iteratively identifying and correcting errors, simplifying intricate steps, and exploring alternative strategies when initial approaches prove inadequate. 
Furthermore, Mulberry \cite{yao2024mulberry} has demonstrated that O1-Like reasoning can also play a significant role in multimodal reasoning. This reasoning paradigm significantly enhances the problem-solving capabilities of large language models (LLMs) by allowing them to approach complex tasks in a more systematic and human-like manner, demonstrating an ability to handle problems that would otherwise be challenging or intractable for conventional LLMs. 

While long-thought reasoning enhances reasoning capabilities and improves accuracy, it is accompanied by longer output sequences, which result in increased computational overhead. A critical challenge lies in developing mechanisms that enable LLMs to dynamically adjust the length and complexity of their reasoning processes in accordance with the difficulty of the problems they encounter.

In this paper, we first revisit the long-thought reasoning processes. we observe that the reasoning processes in long-thought reasoning LLMs often exhibit significant redundancies, which leads to inefficient use of computational resources. This inefficiency not only increases inference costs but also highlights a fundamental limitation in the models' ability to adapt their reasoning depth to suit the demands of diverse tasks. Building on this analysis, we formulate an optimization objective aimed at minimizing reasoning overhead while maintaining accuracy as a constraint. 
Our approach introduces a Length-Harmonizing Reward, which explicitly rewards shorter solutions while penalizing accuracy degradation. By embedding this reward into a RL-based framework, we enable the model to optimize for efficiency without compromising performance. Moreover, our method incorporates an off-policy training strategy inspired by Proximal Policy Optimization (PPO), which aimed at reducing training complexity while maintaining robustness.

Our experiments are conducted using open-source long-thought reasoning LLMs, and we compare our approach against several competing methods like SFT and DPO \cite{rafailov2024directpreferenceoptimizationlanguage}. Through extensive experiments, we demonstrate the efficiency of our proposed methods. Additionally, we perform further studies on the influence of hyperparameters and dataset difficulty on our approach, in order to gain deeper insights into the characteristics and behavior of this novel framework. 

In conclusion, our contributions can be outlined as follows:

\begin{itemize}
    \item We design a simple experiment and identify a critical issue in the reasoning process of long-thought models, referred to as length disharmony, which leads to redundant inference overhead. 
    \item We formulate an optimization problem aimed at improving model inference efficiency while maintaining accuracy, and based on this, we propose Length-Harmonizing Fine-Tuning (\textbf{\method{}}) approach.
    \item Through extensive experiments, we demonstrate the effectiveness of \textbf{\method{}} and conduct in-depth analyses, to provide insights and inspiration for future research in this area.
\end{itemize}

%% file: text/related_work.tex
\section{Related Work}

\textbf{Inference-time Scaling.} 
Inference-time scaling refers to the ability of large language models (LLMs) to improve their outputs by utilizing additional computation during inference time. Recent studies \cite{snell2024scalingllmtesttimecompute} have explored how scaling inference-time computation can enhance the performance of LLMs on challenging prompts. This approach draws parallels to human reasoning, where additional cognitive effort is often allocated to complex problems. In addition to increasing the number of candidate solutions or searching different steps, OpenAI's O1 inference \cite{o12024} demonstrates that extending the length of the solution generated during reasoning can also significantly enhance the model's performance.

\textbf{LLM Alignment.} 
LLM alignment \cite{shen2023largelanguagemodelalignment} constitutes a technical process aimed at guaranteeing that the responses generated by large language models are not only precise and logically consistent but also secure, morally sound, and aligned with the expectations of both developers and users. Ensuring that these expansive language models are in harmony with human preference is crucial for leveraging their immense capabilities in a manner that is both reliable and conscientious. Common methodologies employed in LLM alignment include Supervised Fine-Tuning \cite{zhou2023limaalignment}, Reinforcement Learning from Human Feedback (RLHF) \cite{ouyang2022traininglanguagemodelsfollow}, and Direct Preference Optimization (DPO), among others. The discourse on long thought reasoning optimization presented in this paper can be regarded as an extended setting of LLM alignment, where human preferences are inclined towards shorter outputs (faster inference) and enhanced reasoning accuracy.

\textbf{CoT Compression.} 
Chain-of-Thought (CoT) \cite{wei2023chainofthoughtpromptingelicitsreasoning} and its variations (ToT, \cite{yao2023treethoughtsdeliberateproblem}, GoT \cite{Besta_2024}) are powerful techniques for improving the reasoning capabilities of LLMs. Although CoT is highly effective, it introduces additional computational overhead. Consequently, several studies have attempted to address this issue. For example, \cite{han2024tokenbudgetawarellmreasoning}  introduced a token-budget-aware reasoning framework for large language models (LLMs), which dynamically allocates token budgets according to the complexity of different problems and leverages these budgets to guide the reasoning process. C3oT \cite{kang2024c3otgeneratingshorterchainofthought} employs GPT-4 as a compressor to retain critical information during the reasoning process, thereby reducing reasoning redundancy. Furthermore, several approaches try to utilize continuous representations to mitigate the computational overhead associated with Chain-of-Thought (CoT). For example, CCoT \cite{cheng2024compressedchainthoughtefficient} reduces reasoning overhead by generating contentful and continuous contemplation tokens of variable sequence lengths. COCONUT \cite{hao2024traininglargelanguagemodels}
train LLMs to reason  with fewer thinking tokens during inference in a continuous latent space. However, unlike traditional approaches that focus on compressing normal Chain-of-Thought (CoT), our method centers on long thought reasoning and reduces redundancy in such reasoning by optimizing the reasoning paths instead of compressing each reasoning step. 

Some concurrent works, such as \cite{chen2024think23overthinkingo1like}, have identified the issue of overthinking in O1 reasoning and employs SimPO\cite{meng2024simposimplepreferenceoptimization} for optimization, which is based on the view of preference learning.  And \cite{kimiteam2025kimik15scalingreinforcement} propose long2short RL, using long-CoT techniques to improve short-CoT models. However, in this paper we analyze the long-thought model from a different perspective of length distribution. Moreover, we establish an optimization problem and propose a RL-based method to optimize the model, which provides a different and novel perspective for subsequent research. 



%% file: text/rethinking.tex
\begin{figure*}[!htbp] 
    \centering
    \begin{subfigure}{0.28\textwidth}
        \centering
        \includegraphics[width=\textwidth]{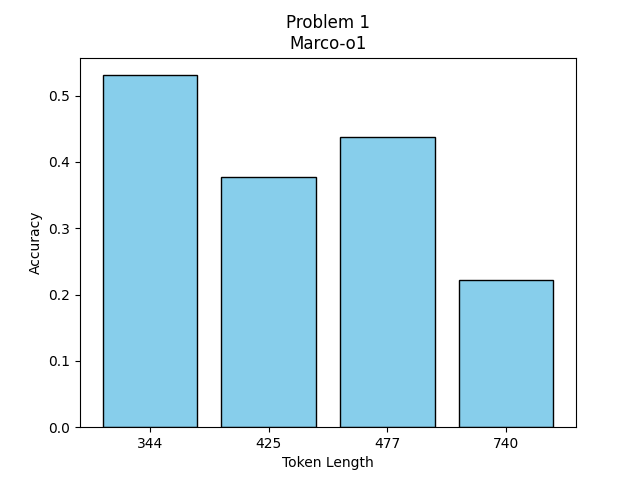}
    \end{subfigure}
    \begin{subfigure}{0.28\textwidth}
        \centering
        \includegraphics[width=\textwidth]{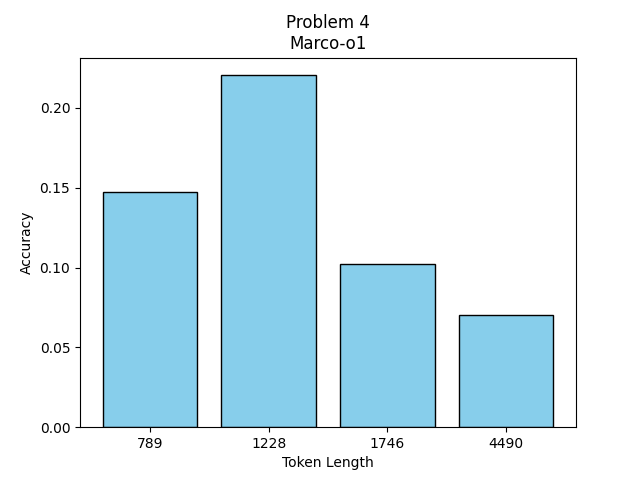}
    \end{subfigure}
    \begin{subfigure}{0.28\textwidth}
        \centering
        \includegraphics[width=\textwidth]{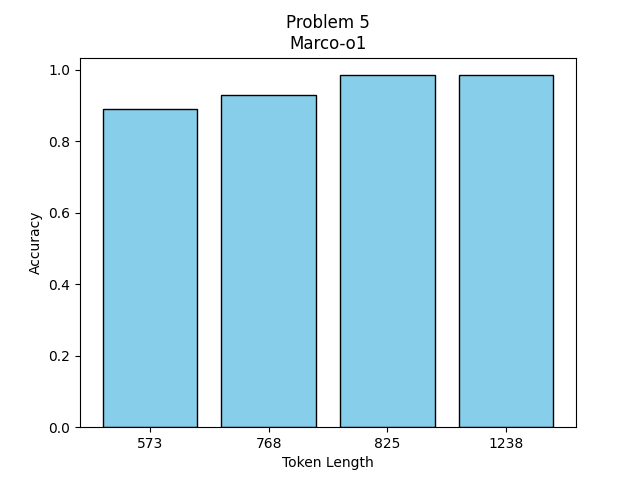}
    \end{subfigure}

    \vspace{2pt} 
    \begin{subfigure}{0.28\textwidth}
        \centering
        \includegraphics[width=\textwidth]{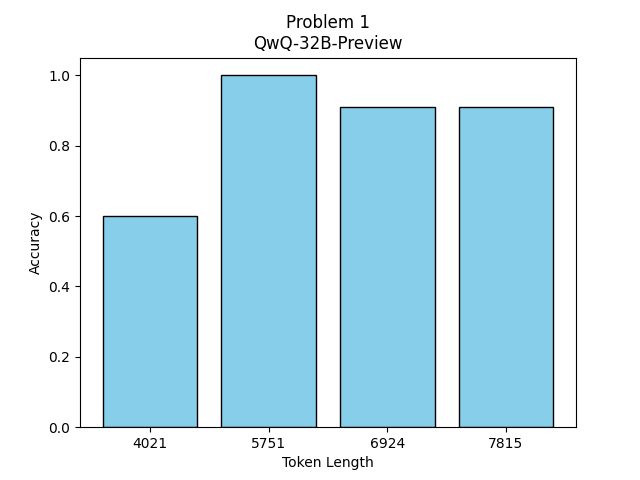}
    \end{subfigure}
    \begin{subfigure}{0.28\textwidth}
        \centering
        \includegraphics[width=\textwidth]{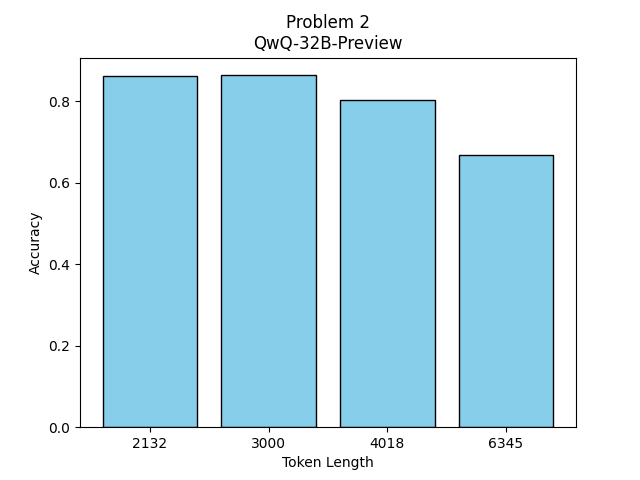}
    \end{subfigure}
    \begin{subfigure}{0.28\textwidth}
        \centering
        \includegraphics[width=\textwidth]{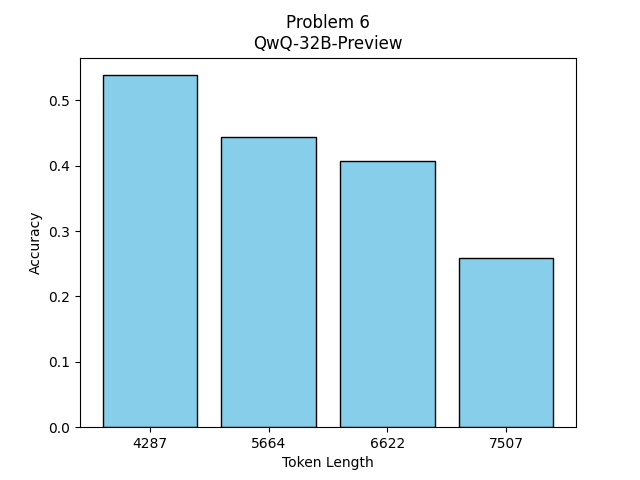}
    \end{subfigure}
    \caption{Accuracy-Length Relationship at Instance level. The relationship between length and accuracy varies significantly across problems, with peak accuracy occurring at short, medium, or long intervals. Notably, high accuracy often persists in shorter intervals.}
    \label{fig:instance-level}
    \vspace{-0.4cm}
\end{figure*}

\section{Revisiting the ``Length Disharmony" in Long Thought Reasoning}
We employ the term ``Length Disharmony" to characterize the phenomenon of inefficiency in the reasoning process of long-thought reasoning, when the model generates responses of varying lengths, among which the shorter responses possess sufficiently high accuracy, thereby rendering the longer responses a superfluous expenditure of computational resources. Besides, due to the quadratic complexity of the Transformer architecture, this will significantly leads to an increase in inference time.

In this section, we have devised a simple experiment to substantiate the disharmony inherent in long thought reasoning. We randomly selected 64 problems from the MATH \cite{hendrycks2021measuringmathematicalproblemsolving} test set (For QwQ-32B, we filtered out hard samples first). For each problem, we generated 512 solutions using both the Marco-o1 and the QwQ-32B models through Top-P sampling \cite{holtzman2020curiouscaseneuraltext}. For each problem, we categorize all candidate solutions into 4 intervals based on their lengths and subsequently compute the accuracy rate for each interval.

\textbf{Accuracy-Length Relationship at Instance Level.} From the data we collected, we can ascertain the relationship between accuracy and length at the instance level, which is shown in Figure \ref{fig:instance-level}. It is evident that there exists a markedly inconsistent relationship between length and accuracy across different problems. The highest accuracy may manifest within the shortest, intermediate, or longest length intervals. Specifically, we observe that relatively high accuracy is preserved even within shorter-length intervals.

\textbf{Accuracy-Length Relationship at Distribution Level.}  Furthermore, by calculating the average accuracy across all problems within different intervals, we have derived the relationship between accuracy and length at the distribution level, which is shown in Table \ref{tab:distribution_level}. At the distribution level, our analysis reveals a consistent trend where shorter response lengths are associated with higher average accuracy rates. This observation can be explained by the premise that a shorter response length typically signifies the model's ability to identify the optimal solution path more efficiently, consequently requiring fewer iterative processes of reflection and backtracking.


Therefore, we can conclude that long-thought models exhibit a phenomenon of length disharmony during reasoning, which leads to redundant computational overhead in the inference phase. This reasoning redundancy can be mitigated, as high accuracy is still maintained even at shorter lengths. From this perspective, we propose Length-Harmonizing Fine-Tuning (\method{}) to optimize long-thought reasoning, enabling it to maintain high accuracy while reducing inference redundancy.

\begin{table}[H]
\centering
\vspace{-0.2cm}
\caption{Accuracy-Length Relationship at Distribution Level. A larger interval number indicates a longer solution length. The average accuracy is higher when the solution length is short.}
\label{tab:distribution_level}
\resizebox{0.98\linewidth}{!}{
\begin{tabular}{lcccc}
\toprule
\textbf{Model} & $\textbf{Interval 1}$ & $\textbf{Interval 2}$ & $\textbf{Interval 3}$& $\textbf{Interval 4}$\\
\midrule
Marco & 81.1 & 80.2 & 78.8 & 75.3 \\
QwQ & 44.9 & 49.9 & 45.9 & 45.3 \\
\midrule
\bottomrule
\end{tabular}
}
\vspace{-0.4cm}
\end{table}

%% file: text/method.tex
\begin{figure*}[ht]
\centering
\includegraphics[width=6.3in]{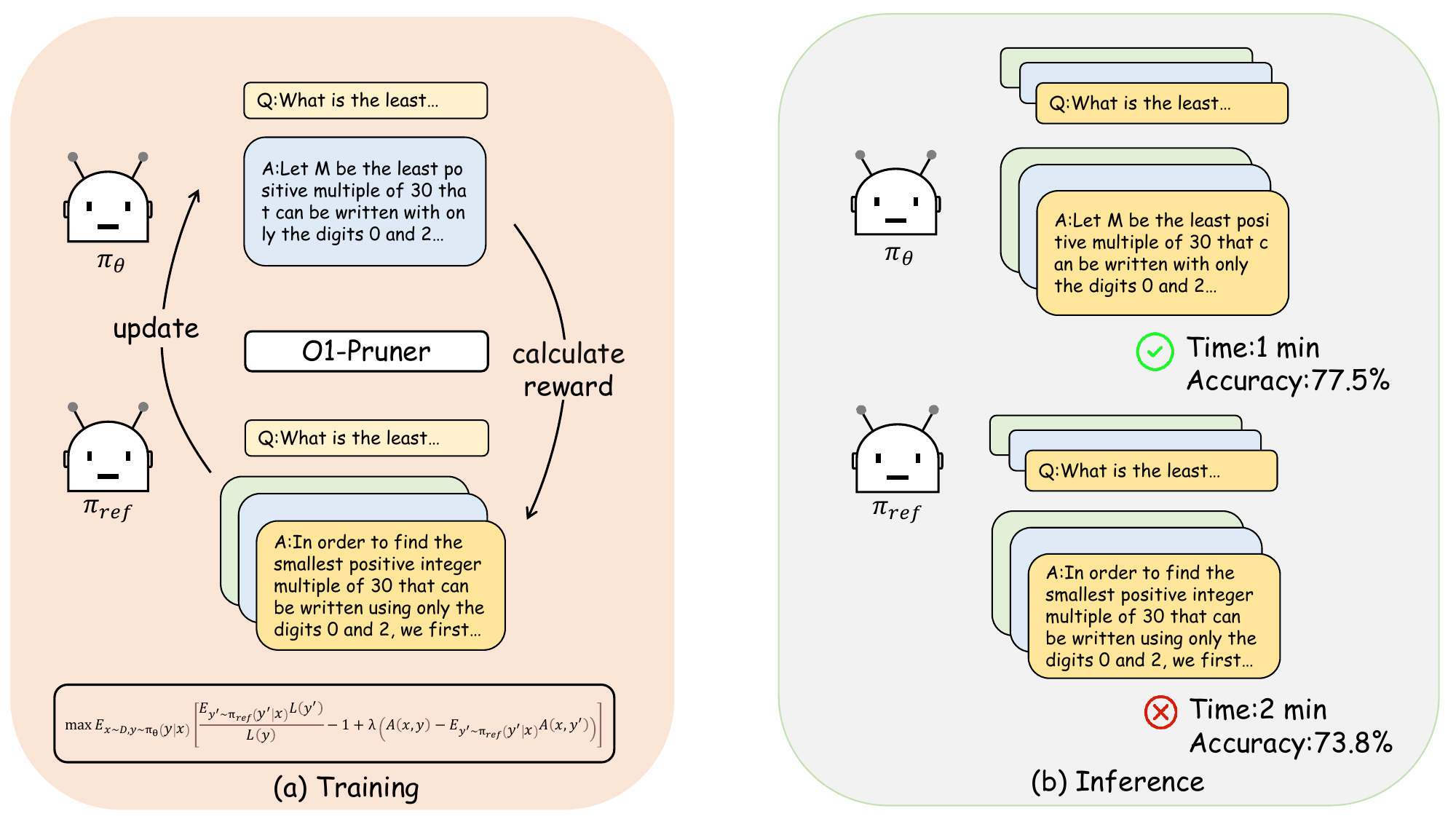}
\caption{Length-Harmonizing Fine-Tuning. During the training phase, for each problem, we sample multiple times from the reference model. Subsequently, we sample from the model to be optimized and compute the reward based on the reference samples, followed by a RL-style fine-tuning. During the inference phase, the model optimized through \textbf{\method{}} demonstrates a significant improvement in inference speed, along with a noticeable enhancement in accuracy.} 
\label{fig:pipeline}
\end{figure*}

\section{Methodology}
In this section, we elaborate on our proposed Length-Harmonizing Fine-Tuning (\method{}) in detail and provide a simple and intuitive mathematical analysis elucidating how our method works for optimize long thought of reasoning.
\subsection{Problem Setup}
We consider a LLM parameterized by $\mathbf{\theta}$ and denoted as $\pi_{\mathbf{\theta}}$. In the context of math problem solving, the LLM  accepts a sequence $\mathbf{x} = [x_1, \ldots, x_n]$, commonly termed as the problem, and then generate a corresponding solution $\mathbf{y} = [y_1, \ldots, y_m]$. Hence, the solution $\mathbf{y}$ is construed as a sample drawn from the conditional probability distribution $\pi_{\mathbf{\theta}}(\cdot | \mathbf{x})$. The conditional probability distribution $\pi_{\mathbf{\theta}}(\mathbf{y} | \mathbf{x})$ can be decomposed as follows:
\begin{align}
\pi_{\mathbf{\theta}}(\mathbf{y} | \mathbf{x}) = \prod_{j=1}^{m} \pi_{\mathbf{\theta}}(y_{j} | \mathbf{x}, \mathbf{y}_{<j}).
\end{align}
Firstly, we review the process of supervised fine-tuning (SFT). SFT is the primary method to adapt a pre-trained LLM for downstream tasks with a relatively smaller supervised dataset of labeled examples compared to the data of pre-training stage.
In this paper, we focus on the task of mathematic problem solving where the problem-solution pairs denoted as $(\mathbf{x},\mathbf{y})$, are drawn from a specified SFT dataset $\mathcal{D}$. Thus the training objective of SFT under this setting can be formulated as:
\begin{align}
\max_{\pi_{\theta}}  \mathbb{E}_{(x,y)\sim \mathcal{D}} 
 \Big[\log \pi_{\mathbf{\theta}}(\mathbf{y}\ |\  \mathbf{x})\Big].
\end{align}

\subsection{Length-Harmonizing Fine-Tuning (\method{})}
To start with, let's assume that $\pi_{\theta}$ is a LLM that can solve math problems with long thought with redundancy and disharmony. 
we hypothesize that the reasoning paths represented by output thought of language model $\pi_{\theta}$ contain redundancies and lack proper coordination. To address this, we propose an optimization objective that ensures no degradation in accuracy while tackling the issue from two perspectives. First, at the overall level, we aim to shorten the reasoning paths. Second, we encourage the model to output shorter answers for simpler problems, while for more complex problems, we guide the model to learn the correct reasoning paths, which, according to the inference scaling law, typically involve longer reasoning sequences. Given a problem \(x\), we define \(L(y)\) as the length (counted by token) of the solution \(y\). Considering a reference model $\pi_{ref}$, we reduce the solution length of the policy model relative to that of the reference model, which can be formulated as:
\begin{align}
\max \mathbb{E}_{x \sim D} \left[ \mathbb{E}_{y \sim \pi_{\theta}(y | x), y' \sim \pi_{ref}(y | x)}
\frac{L(y')}{L(y)}  - 1 \right].
\end{align}
We subtract a constant 1 from the optimization objective to ensure that the initial expected value of the optimization is zero. We then define an accuracy function \(A(x, y, \text{answer})\), which takes the problem, solution, and the real answer as inputs, and returns 0 or 1 to indicate whether the solution is incorrect or correct. For the sake of simplicity in the notation, we omit the real answer, denoting the function as \(A(x, y)\). We aim to ensure that the model's accuracy does not decrease, or even improves, during the process of optimizing for length. Thus, we derive the following constraint condition:
\begin{align}
\!\!\! \mathbb{E}_{x \sim D, y \sim \pi_{\theta}(y | x)} A(x, y) \geq 
\mathbb{E}_{x \sim D, y' \sim \pi_{ref}(y' | x)} A(x, y').
\end{align}
Therefore, we can establish our optimization objective as: 
\begin{align}
    &\max \mathbb{E}_{x \sim D} \left[ \mathbb{E}_{y \sim \pi_{\theta}(y | x), y' \sim \pi_{ref}(y | x)}
    \frac{L(y')}{L(y)}  - 1 \right] \\ \text{s.t.}\ 
    & \mathbb{E}_{x \sim D, y \sim \pi_{\theta}(y | x)} A(x, y) \geq 
    \mathbb{E}_{x \sim D, y' \sim \pi_{ref}(y' | x)} A(x, y'). \notag
\end{align}
To solve this constrained optimization problem, we incorporate constraint into the objective function as a penalty term. Specifically, the constraint on accuracy is added to the objective with a penalty weight $\lambda \geq 0$: 
\begin{align}
\max \mathbb{E}_{x \sim D,y \sim \pi_{\theta}(y | x), y' \sim \pi_{ref}(y | x)} \frac{L(y')}{L(y)} - 1 \notag\\
+ \lambda (A(x, y) - A(x, y')).
\end{align}
By reorganizing the terms related with reference model $\pi_{ref}$, we have:
\begin{align}
\max \mathbb{E}_{x \sim D,y \sim \pi_{\theta}(y | x)}     \frac{\mathbb{E}_{y' \sim \pi_{ref}(y' | x)} L(y')}{L(y)}  - 1 + \notag\\  \lambda (A(x, y) - \mathbb{E}_{y' \sim \pi_{ref}(y' | x)} A(x, y')).
\end{align}
In practice, we approximate the expectation terms related with $\pi_{ref}$ by sampling. For each $x$, we sample for $K$ times from $\pi_{ref}(\cdot | x)$ and calculate the mean value:
\begin{gather}
    \bar{L}_{ref}(x) = \frac{1}{K} \sum_{i=1}^K L(y'_i),\quad y'_i \sim \pi_{ref}(\cdot \mid x); \\
    \bar{A}_{ref}(x) = \frac{1}{K} \sum_{i=1}^K A(x,y'_{i}),\quad y'_i \sim \pi_{ref}(\cdot \mid x);
\end{gather}
This approach is widely employed in Policy Gradient with Baseline. Furthermore, a recently proposed method GRPO \cite{shao2024deepseekmathpushinglimitsmathematical} adopts a similar technique to reduce training overhead. Based on this technique, our objective can be approximated as:
\begin{align}
\max \mathbb{E}_{x \sim D, y \sim \pi_{\theta}(y | x)}     \frac{\bar{L}_{ref}(x)}{L(y)}  - 1 \notag\\ +   \lambda (A(x, y) - \bar{A}_{ref}(x)). 
\end{align}
Since both $L(y)$ and $A(x,y)$ are not differentiable, we solving this optimization with policy gradient approach, which is shown to have strong performance despite its simplicity. Furthermore, it is worth noting that during the optimization process, frequent sampling from the current distribution \( \pi_\theta \) is required during training, which significantly increases the complexity of the training procedure. Considering that off-policy training can bring remarkable effectiveness with pre-collected data, we adopt an off-policy training approach by directly sampling from the \( \pi_{ref} \) instead of \( \pi_\theta \). Besides, since our reward is derived by assessing the merit of a sample within the distribution relative to the expected outcome, our reward can be regarded as an approximate advantage function. Consequently, we employ a PPO-style loss \cite{schulman2017proximalpolicyoptimizationalgorithms} to optimize the objective function, which helps for our off-policy training strategy. Defining the Length-Harmonizing Reward $R_{LH}(x, y) = \frac{\bar{L}_{ref}(x)}{L(y)}  - 1 \notag\\ +   \lambda (A(x, y) - \bar{A}_{ref}(x))$, the loss function of off-policy-version Length-Harmonizing Fine-Tuning is:
\begin{align}
\!\!\!
& L^{\text{LH}}(\theta;x, y) 
= -\mathbb{E}_{x \sim D,y \sim \pi_{ref}(y | x)} \big[ \min ( r(\theta) R_{LH}(x, y), \,\notag\\ 
&  \qquad \qquad \text{clip}(r(\theta), 1 - \epsilon, 1 + \epsilon) R_{LH}(x, y) )\big],
\end{align}
where $r(\theta) = \frac{\pi_{\mathbf{\theta}}(\mathbf{y} | \mathbf{x})}{\pi_{ref}(\mathbf{y} | \mathbf{x})}$. clip() is the clipping function.

This allows us to prepare the required data at the beginning of training, thereby greatly simplifying the training workflow. Our experiments show that this off-policy approach still enables our method to achieve outstanding performance, significantly surpassing other baselines.



\subsection{Understanding the Loss Function}
To intuitively understand how our loss function works, we begin by analyzing the $R_{LH}$ term. Evidently, $R_{LH}$ comprises two distinct components, namely the length reward term $\frac{\bar{L}(x,\pi_{ref})}{L(y)}  - 1$ and the accuracy reward term $\lambda (A(x, y) - \bar{A}(x,\pi_{ref}))$. Obviously, the length reward term will reward shorter outputs. When the sequence length are consistent with expected output length of reference model, the length reward is 0; however, when the output is longer, the length reward becomes negative. The accuracy reward term is essential for balancing length and accuracy. For a problem \(x\) with a relatively high accuracy expectation, solving it correctly does not yield a significant accuracy reward. As a result, the model tends to explore shorter solutions. For more challenging problems, solving them correctly yields a higher accuracy reward, indicating that we do not want the model to prioritize shortening the output. Instead, we aim for the model to focus on generating a correct solution. On this basis, if the correct solution is relatively short, the model will receive an additional length reward.

To the end, we summarize the training procedure of our proposed \method{} in Algorithm \ref{alg:o1-pruner}.

\begin{algorithm}[t]
    \caption{O1-Pruner}
    \label{alg:o1-pruner}
\begin{algorithmic}[1]
    \STATE {\bfseries Input:} LLM $\pi_\theta$, Dataset $\mathcal{D} = \{(x^i, a^i)\}_{i \in [N]}$
    \STATE {\bfseries Initialize:} $\pi_{ref}=\pi_\theta$
    \FOR{$i=1$ {\bfseries to} $N$} 
        \STATE sampling K solutions $y^i_{1}$, ...,$y^i_{K}$ from $\pi_{ref}(\cdot|x_i)$
        \STATE calculating $\bar{L}_{ref}(x^i) = \frac{1}{K}\sum_{k=1}^K L(y^i_k)$
        \STATE calculating $\bar{A}_{ref}(x^i) = \frac{1}{K} \sum_{k=1}^K A(x^i,y^i_k)$
        \STATE randomly select m ($m\leq K$) solutions from $y^i_{1}$, ...,$y^i_{K}$
        \STATE Update $\theta = \mathop{\arg\min}\limits_{\theta} \sum_{j=1}^m L^{\text{LH}}(\theta;x^i, y^i_{j})$ 
    \ENDFOR
    \STATE {\bfseries Output:} Updated LLM $\pi_\theta$ 
    
    

\end{algorithmic}
\end{algorithm}

%% file: text/experiments.tex
\section{Experiments}

In this section, we conduct extensive experiments to verify the efficacy of the proposed \method{}.

\begin{table*}[t]	
	\begin{center}
        \caption{Main Experiment Results. We present the performance of two selected models optimized through different methods across three mathematical reasoning datasets. It can be observed that the models trained with \textbf{\method{}} achieve the best trade off between accuracy and length in comparison with other approaches.} 
        \label{tab:main_results}
\resizebox{0.95\linewidth}{!}{
			\begin{tabular}{lcccccccccccc}
				\hline\noalign{\smallskip}
				\multirow{3}*{Model} & \multicolumn{3}{c}{MATH} &\multicolumn{3}{c}{GSM8K} &\multicolumn{3}{c}{GaoKao} &\multicolumn{3}{c}{\textit{AVERAGE}}\\
				\cmidrule(r){2-4}\cmidrule(r){5-7}\cmidrule(r){8-10}\cmidrule(r){11-13}
				& Acc & Length & AES & Acc & Length & AES & Acc & Length & AES & Acc & Length & AES\\
				\noalign{\smallskip}
				\noalign{\smallskip}\hline\noalign{\smallskip}
		\noalign{\smallskip}\hline
        \textit{\textbf{Marco-o1-7B}}\\
        \textit{(full fine-tune)}\\
        Baseline & \underline{73.8} & 1156 & 0 & 89.2 & 530 & 0 & 57.1 & 1112 & 0 & \underline{73.4} & 932 & 0\\
        Fast-solving Prompt & 71.0 & 1113 & 0.15 & 81.7 & 447 & \underline{0.41} & \underline{57.1} & 1062 & 0.04 & 69.9 & 874 & 0.20\\
        SFT & 73.6 & 1076 & 0.08 & \underline{89.9} & 497 & 0.09 & 56.3 & 1066 & 0.08 & 73.3 & 880 & 0.08\\
        DPO & 71.8 & \underline{761} & \underline{0.42} & 88.6 & \underline{410} & 0.25 & 56.6 & \underline{780} & \underline{0.32} & 72.3 & \underline{650} & \underline{0.33}\\

        \method{} & \underline{\textbf{77.5}} & \underline{\textbf{657}} & \underline{\textbf{0.58}} & \underline{\textbf{91.4}} & \underline{\textbf{343}} & \underline{\textbf{0.43}} & \underline{\textbf{61.6}} & \underline{\textbf{664}} & \underline{\textbf{0.64}} & \underline{\textbf{76.8}} & \underline{\textbf{554}} & \underline{\textbf{0.55}}\\
        \noalign{\smallskip}\hline\noalign{\smallskip}
            \textit{\textbf{QwQ-32B-Preview}}\\
        \textit{(freeze fine-tune last 48 layers)}\\
        Baseline & 90.6 & 2191 & 0 & 95.1 & 777 & 0 & 79.0 & 2183 & 0 & 88.2 & 1717 & 0\\
        Fast-solving Prompt & 90.2 & \underline{1763} & \underline{0.21} & \underline{95.8} & \underline{561} & \underline{0.30} & 78.4 & \underline{1911} & \underline{0.15} & 88.1 & \underline{1411} & \underline{0.22}\\
        SFT & 90.4 & 2031 & 0.08 & 95.7 & 717 & 0.10 & 79.5 & 2112 & 0.05 & 88.5 & 1620 & 0.08\\
        DPO & \underline{\textbf{91.7}} & 1999 & 0.12 & 95.3 & 704 & 0.10 & \underline{79.7} & 2021 & 0.10 & \underline{88.9} & 1575 & 0.11\\
        \method{} & \underline{91.0} & \underline{\textbf{1385}} & \underline{\textbf{0.38}} & \underline{\textbf{96.5}} & \underline{\textbf{534}} & \underline{\textbf{0.36}} & \underline{\textbf{80.3}} & \underline{\textbf{1446}} & \underline{\textbf{0.39}} & \underline{\textbf{89.3}} & \underline{\textbf{1121}} & \underline{\textbf{0.38}}\\
        \noalign{\smallskip}\hline
			\end{tabular}
		}
	\end{center}
\end{table*}

\subsection{Experiment Setup}

\textbf{Long-thought Models.}  The long thought models we chosen for our experiment are Marco-o1-7B and QwQ-32B-Preview, which have demonstrated excellent performance on a wide range of math problem-solving tasks. For Marco-o1-7B, we utilize full-parameter fine-tuning; however, for the larger-scale QwQ-32B-Preview, our computational resources are not able to support full-parameter training. As a result, we adopt Parameter-Efficient Fine-Tuning \cite{han2024parameterefficientfinetuninglargemodels}. After evaluating both LoRA \cite{hu2021loralowrankadaptationlarge} and Freeze Fine-Tune, we observed that Freeze Fine-Tune yields much better performance. Therefore, we selected this fine-tuning approach for our experiments.

\textbf{Dataset.}
The dataset used for training is MATH. It comprises approximately 10k math problem of high school level accompanied with both ground truth solution and ground truth answer. Since the ground truth solution is not need for our experiment, we only use the problem-answer pairs. For training, we selected 5,000 problems from the MATH Trainset. For Marco-o1-7B, we generated 16 solutions for each problem; for QwQ-32B-Preview, we generated 12 solutions for each problem. The dataset utilized for testing encompasses the test sets of MATH, GSM8k \cite{cobbe2021trainingverifierssolvemath}, and GaoKao (mathematical) \cite{zhang2024evaluatingperformancelargelanguage}, comprising a diverse range of mathematical problems with varying levels of difficulty.

\textbf{Baselines.} 
To validate the superiority of our method for long thought reasoning optimization tasks, we have selected the three representatively comparative methods. \textbf{(i) Fast-Solving Prompt:} The Fast-Solving Prompt is a prompting technique wherein we instruct the model within the prompt to solve the given problem as swiftly as possible, aiming to achieve the desired reduction in reasoning length.  \textbf{(ii) SFT:} For the SFT method, we curated the training dataset by selecting the two shortest correct solutions for each problem, ensuring that the model is exposed to examples that embody both accuracy and conciseness. These solutions were then used to train the model following the standard SFT pipeline. \textbf{(iii) DPO:} For the implementation of DPO, we meticulously selected two of the shortest correct solutions to serve as the chosen samples, which exemplify efficiency and precision in problem-solving. Conversely, to represent the reject sample, we opted for the longest solution available.

\textbf{Evaluation Metric.} 
We employ the following average accuracy, average length and Accuracy-Efficiency Score (AES) as key metrics to assess whether the model achieves a desirable balance between reasoning accuracy and  length:
\begin{itemize}
    \item \textbf{Accuracy} Accuracy reflects whether the model correctly solves the problem. It is measured as the proportion of problems for which the model's generated solution is correct. A higher accuracy indicates better problem-solving capability.
    \item \textbf{Length} Length denotes the number of tokens in the generated solution. It serves as a proxy for the computational cost of generating solutions, where a shorter length implies greater efficiency.
    \item \textbf{AES} We define a novel metric called  Accuracy-Efficiency Score (AES), to evaluate the trade off between improving accuracy and reducing computational cost. It is calculated by weighting and summing the model's solution length and accuracy. Defining $\Delta \text{Length} = \frac{\text{Length}_{\text{baseline}}-\text{Length}_{\text{model}}}{\text{Length}_{\text{baseline}}}$ and 
    $\Delta \text{Acc} = \frac{\text{Acc}_{\text{model}}-\text{Acc}_{\text{baseline}}}{\text{Acc}_{\text{baseline}}}$, the AES is calculated by:
    \begin{equation*}
    \text{AES} =
    \begin{cases} 
    \alpha \cdot \Delta \text{Length} + \beta \cdot \left\vert\Delta \text{Acc}\right\vert, & \text{if } \Delta \text{Acc} \geq 0 \\
    \alpha \cdot \Delta \text{Length} - \gamma \cdot \left\vert\Delta \text{Acc}\right\vert, & \text{if } \Delta \text{Acc} < 0
    \end{cases}
    \end{equation*}
    where $\alpha > 0$, $\beta > 0$, and $\gamma > 0$. AES evaluates the trade-off between improving accuracy and reducing computational cost. And we emphasize the penalization of accuracy degradation by setting $\gamma$ $>$ $\beta$. We set the default values as $\alpha=1$, $\beta=3$, $\gamma=5$.
\end{itemize}



\begin{figure*}[htbp]
\centering
\includegraphics[width=6.15in]{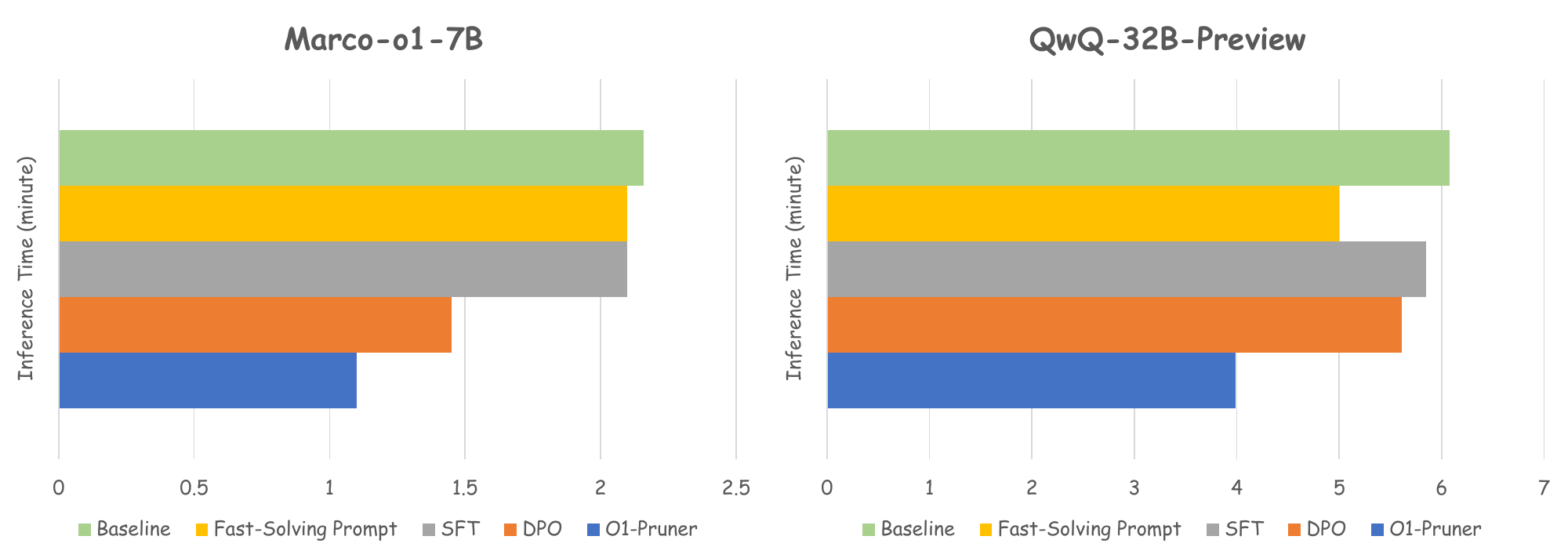}
\vspace{-0.2cm}
\caption{Comparison of inference time-cost on MATH among different models and methods. \textbf{\method{}} achieves the shortest inference times (slightly over 1 minute for Marco-o1-7B and ~4 minutes for QwQ-32B-Preview), demonstrating its effectiveness in accelerating long-thought model inference for both small and large long thought models.} 
\label{fig:timecost}
\end{figure*}

\subsection{Experimental Results}
Table~\ref{tab:main_results} demonstrates the performance of various methods across different evaluation metrics. The proposed \textbf{\method{}} consistently achieves superior performance in balancing reasoning accuracy and efficiency compared to baseline and competing methods. Notably, it exhibits the best trade-off between accuracy and reasoning length across all datasets, as further supported by its significantly higher Accuracy-Efficiency Score (AES) values. 
Across both models, Marco-o1-7B and QwQ-32B-Preview, \textbf{\method{}} outperforms other methods in average length of generated solutions, with a noticeable improvement on accuracy. For instance, in the Marco-o1-7B experiments, \textbf{\method{}} achieves an average accuracy of 76.8\%, accompanied by a 40.5\% reduction in solution length compared to the baseline. Similarly, for QwQ-32B-Preview, \textbf{\method{}} yields an average accuracy of 89.3\%, with a 34.7\% reduction in solution length. These improvements highlight the robustness of \textbf{\method{}} in enhancing computational efficiency without sacrificing accuracy.

The Fast-Solving Prompt method, while achieving a moderate reduction in solution length, compromises accuracy in most cases. This trade-off is evident from its lower AES values compared to \textbf{\method{}}, indicating that the reduction in reasoning length often comes at the cost of problem-solving performance. On the other hand, SFT provides a better balance than the Fast-Solving Prompt, but its improvements in reasoning length remain marginal, with limited gains in AES.
The DPO method achieves a reasonable balance between accuracy and length, but it falls short of the performance achieved by \textbf{\method{}}. Besides, the average accuracy decreases notably on Marco-o1-7B.

\subsection{Inference Time-Cost Analysis}
In this subsection, we take the MATH test set as an example to explore the time overhead during the model inference phase. We utilize one A800 GPU and the VLLM \cite{kwon2023efficientmemorymanagementlarge} library for inference, recording the average inference time. For the Marco-o1 model, we employ one A800 GPU, while for the QwQ-32B-Preview model, we use four A800 GPUs. As illustrated in Figure \ref{fig:timecost}, the inference time results reveal notable differences across methods and models:
For the Marco-o1-7B model, the baseline approach demonstrates an inference time of approximately 2 minutes, while the Fast-Solving Prompt and SFT methods achieve slightly shorter times. Both the DPO and \textbf{\method{}} methods exhibit significantly reduced inference times, with \textbf{\method{}} achieving the shortest duration, slightly exceeding 1 minute.
For the larger model QwQ-32B-Preview, the overall inference time is considerably higher. The Baseline approach records the longest inference time, approaching 6 minutes, while the DPO and SFT methods achieve slightly shorter durations. Notably, the Fast-Solving Prompt reduces the inference time to around 5 minutes, likely due to the strong instruction-following capabilities of large models. Once again, \textbf{\method{}} demonstrates the shortest duration, achieving an inference time of approximately 4 minutes.

In summary, \textbf{\method{}} represents a significant advancement in optimizing long-thought reasoning for math problem-solving tasks for both smaller and larger language models, achieving the best balance between accuracy and efficiency while minimizing computational overhead. 

%% file: text/further_evaluation.tex
\begin{table}[t]
\centering
\caption{Ablation experiments on $\lambda$. Overall, the model's accuracy and solution length increase with the penalty coefficient $\lambda$. A larger $\lambda$ implies that the model places greater emphasis on variations in accuracy, thereby partially weakening the optimization for sequence length. $\lambda$ = 2 achieves an optimal balance between accuracy and efficiency.}
\label{tab:ablation}
\resizebox{0.98\linewidth}{!}{
\begin{tabular}{lccc}
\toprule
$\mathbf{\lambda}$ & $\textbf{Acc}_{avg}$ & $\textbf{Length}_{avg}$ & $\textbf{AES}_{avg}$\\
\midrule
\textit{\textbf{Marco-o1-7B}}\\
0 & 74.8 & 527 & 0.49\\
1 & 76.0 & 532 & 0.54\\
2 & 76.8 & 554 & 0.55\\
5 & 76.3 & 656 & 0.45\\
\midrule
\bottomrule
\end{tabular}
}
\vspace{-0.4cm}
\end{table}

\subsection{Ablation Study}
\textbf{Ablation on Hyper-parameter Sensitivity.} In this part, we evaluate the hyperparameter sensitively of constraint coefficient $\lambda$. We select several different values of $\lambda$ ($\lambda = 0, 1, 2, 5$) and evaluate the model accordingly. For the sake of brevity, we only report the average metrics across three datasets. It can be observed that, overall, the model's accuracy increases as the penalty coefficient lambda rises, while the required inference length also grows. In our experiments, for Marco-o1-7b, setting $\lambda = 2$ achieves a favorable trade-off between accuracy and efficiency.
\begin{figure}[htbp]
\vspace{-0.2cm}
\centering
\includegraphics[width=0.5\textwidth]{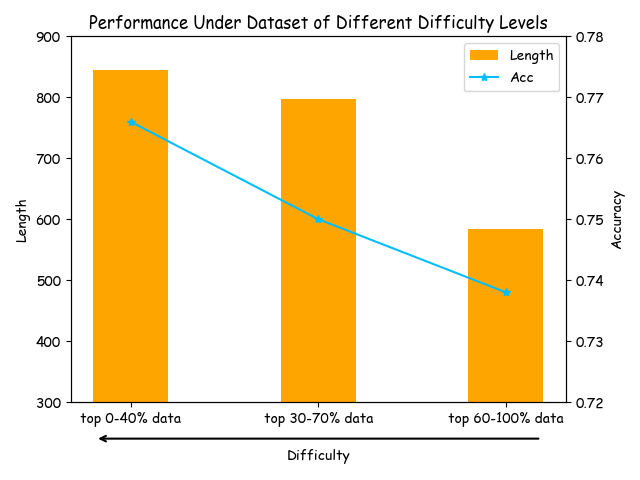}
\vspace{-0.4cm}
\caption{ Performance on MATH Test-set When Trained on Problems of Different Difficulty Levels. Models trained on more challenging datasets tend to generate longer solutions, while learning to solve harder problems enhances model accuracy. In contrast, for less challenging datasets, shorter solutions are produced without a corresponding accuracy improvement.}
\label{fig:difficult}
\vspace{-0.2cm}
\end{figure}

\textbf{Ablation on Difficulty Levels.} We investigate the performance and characteristics of \method{} across datasets of varying difficulty levels. Due to limited computational resources, we exclusively selected Marco-o1 for experimentation. Utilizing the data constructed from the MATH dataset as mentioned in prior experiments (comprising 5k problems * 16 solutions), we partition the dataset into three subsets of differing difficulty based on the model's average accuracy.
In Figure \ref{fig:difficult}, We observe that models trained on more challenging datasets tend to generate longer solutions, as these datasets typically contain problems requiring more complex solutions. At the same time, by learning the correct solutions of harder problems, the models improve their problem-solving capabilities and ultimately achieve higher accuracy. In contrast, for the least challenging datasets, although the generated solution lengths are reduced, there is no improvement in accuracy. These experimental results suggest that while our approach demonstrates significant effectiveness in optimizing long-thought reasoning, it remains highly influenced by the nature of the training data.